\RequirePackage{fix-cm}
\documentclass[smallextended]{svjour3}       
\smartqed  
\usepackage{graphicx}
\usepackage[ruled,vlined,linesnumbered]{algorithm2e}

\usepackage{multirow}
\usepackage{booktabs}
\usepackage{hyperref}
\usepackage{geometry}
\usepackage{amsmath}
\usepackage{amsfonts}
\usepackage{amssymb}
\usepackage{bm}
\usepackage{calc}
\usepackage{subcaption}
\usepackage{xcolor}
\usepackage{mathtools}

\AtBeginDocument{%
  \providecommand\BibTeX{{%
    \normalfont B\kern-0.5em{\scshape i\kern-0.25em b}\kern-0.8em\TeX}}}
\begin{document}

\title{Boosting House Price Predictions using Geo-Spatial Network Embedding
}


\author{Sarkar Snigdha Sarathi Das \and
        Mohammed~Eunus~Ali \and
        Yuan-Fang Li \and
        Yong-Bin~Kang \and Timos Sellis 
}


\institute{Sarkar Snigdha Sarathi Das \and Mohammed Eunus Ali \at
              Bangladesh University of Engineering and Technology \\
              \email{sarathismg, mohammed.eunus.ali@gmail.com}           
           \and
           Yuan-Fang Li \at
                Monash University
              	\email{yuanfang.li@monash.edu}
            \and
            Yong-Bin Kang \and Timos Sellis \at
                Swinburne University of Technology
                \email{ykang, tsellis@swin.edu.au}
}


\maketitle

\begin{abstract}

Real estate contributes significantly to all major economies around the world. 
In particular, house prices have a direct impact on stakeholders, ranging from house buyers to financing companies.
Thus, a plethora of techniques have been developed for real estate price prediction. 
Most of the existing techniques rely on different house features to build a variety of prediction models to predict house prices. Perceiving the effect of spatial dependence on house prices, some later works focused on introducing spatial regression models for improving prediction performance. However, they fail to take into account the geo-spatial context of the neighborhood amenities such as how close a house is to a train station, or a highly-ranked school, or a shopping center. 
Such contextual information may play a vital role in users' interests in a house and thereby has a direct influence on its price. 
In this paper, we propose to leverage the concept of graph neural networks to capture the geo-spatial context of the neighborhood of a house. 
In particular, we present a novel method, the Geo-Spatial Network Embedding (GSNE), that learns the embeddings of houses and various types of Points of Interest (POIs) in the form of multipartite networks, where the houses and the POIs are represented as attributed nodes and the relationships between them as edges. Extensive experiments with a large number of regression techniques show that the embeddings produced by our proposed GSNE technique consistently and significantly improve the performance of the house price prediction task regardless of the downstream regression model. 

\end{abstract}

\section{Introduction}

The price of a house is one of the most critical factors in the decision-making process of buying a house. 
Determining which house to buy is a challenging task as it is influenced by a multitude of other factors: features of houses as well as the complex geo-spatial relationships with their neighborhoods. 

Thus, the task of house price prediction has received significant attention in both academia and stakeholders for decades. 
Over the years, researchers have used different techniques to build effective models for house price predictions. 
For example, typical Hedonic Price models~\cite{hedonic_review,rosen1974hedonic} have been extensively studied to model the relationship between prices and housing features. While the early models based on ordinary least square (OLS)  context did not include spatial awareness, researchers gradually realized the impact of regional submarkets in house price prediction. To incorporate the locational influence, several works \cite{bourassa2003housing,fik2003modeling,dubin1998predicting,bourassa2007spatial} used spatial statistical methods since simple hedonic models are not much effective in handling spatial dependence in regression residuals. However, these spatial statistical methods require explicit feature engineering by domain experts.
Following the success of machine learning in different fields, in recent years, SVM~\cite{wang2014real}, Convolutional Neural Networks~\cite{piao2019housing}, and Recurrent Neural Networks~\cite{chen2017house} have been employed to better capture the preferences for more accurate house price prediction. House images have also been leveraged for better price estimation~\cite{zhao2019deep}. Unlike OLS based hedonic models, these modern learning algorithms can effectively capture the spatial dependence from location attributes which obviates the need of explicit feature engineering.

Although most of the recent prior works leverage the detailed housing and location features, they overlook the geo-spatial contexts such as ``\textsf{how close is this house to the train station?}'', or ``\textsf{is there any good school in walking distance from the house?}''. 
These geo-spatial contexts based on neighbourhood facilities can greatly influence user preferences on buying a house and hence can be key determining factors for the price of the house. 
To illustrate, suppose that there are two houses with 
the same set of features such as the number of bedrooms, house areas, etc., in the same suburb. However, there may have variations in their prices. 
For example, a house next to the train station will likely have a much higher price than the house which is far away (e.g.\ 3km distance) from the train station. 
Similarly, houses in a certain suburb with good schools in the neighborhood and a train station for commuting to the city will have higher prices than those houses in a nearby suburb that does not have a good school and a train station in its neighborhood. 
Hence, points of interests (POIs) such as train stations, schools, shopping centers, etc.\ in the neighborhood can play key roles in determining the house prices. 
To the best of our knowledge, none of the prior works capture the important features related to neighborhood POIs and their relationship with houses.

It is not straightforward to capture complex latent interactions between houses and POIs as it involves connectivity among different entities (e.g.\ how close a house is to the train station) as well as heterogeneous sets of features of these entities (e.g.\ how good a school is).
 Recently, an approach by Jenkins et. al.~\cite{jenkins2019unsupervised} used satellite image, taxi mobility data, and the existence of different categories of point of interests for generating an embedding for a region, which are later used to a get a coarse outline of price distribution per sqft for houses in that region. Though this work can capture the regional features using complex sets of data, they neither consider detailed houses and POIs features nor the relationships between neighborhood POIs and the corresponding house, which is our main focus in this paper.

In this paper, we propose Geo-Spatial Network Embedding (GSNE) that accurately captures highly useful spatial features (both connectivity and the content) of key neighborhood POIs such as schools, train stations, etc. and their relations with the houses. 
We leverage the key concept of graph embedding that essentially learns low dimensional feature representations of a given attributed graph. GSNE employs Gaussian-based embedding methods~\cite{bojchevski2017deep,hettige2020gaussian,zhu2018deep} as they have been shown to be effective and robust against noises and uncertainties that are inherent in many real-world graphs such as house-neighborhood networks in real estate. 

We propose to represent houses and their neighborhood POIs as a multipartite graph, in which nodes of different partitions (types) represent houses and POIs (e.g.\ regions and schools), and edges represent the spatial proximity relation between nodes. 
In our case, the graph is attributed and weighted, where nodes are attributed with their own features and edge weights represent the distance between two nodes. 
The key intuition of our proposed approach, GSNE, is to project the nodes in a \emph{spatial network} into a Gaussian feature space. 

Prior works in Gaussian-based network embeddings~\cite{bojchevski2017deep,hettige2020gaussian,zhu2018deep} only consider homogeneous or bipartite network, and it is not straightforward to embed heterogeneous multipartite networks in the same Gaussian space. The heterogeneity poses several challenges that include projecting different categories of node features into the same Gaussian space and developing effective sampling strategies and training schemes. We address all of these challenges in our proposed GSNE framework.

We evaluate our GSNE framework on the house price prediction task on a large real-estate and POI datasets of Melbourne, Australia. 
We concatenate the learned embedded vectors from GSNE, which capture essential spatial information about houses and their neighborhood context, with the raw house features vectors as features to predict house prices. 
Compared with raw features only, the concatenated features achieve the best prediction performance on a large number of regression models, demonstrating the effectiveness and robustness of our GSNE model. 

In summary, our contributions are as follows:
\begin{itemize}

\item We propose a novel geo-spatial network embedding (GSNE) framework that can accurately capture the geo-spatial neighborhood context in terms of different types of POI and their features, and the relationships among these POIs in a weighted, attributed multipartite graph.

\item We adopt and extend the Gaussian embedding methods to realize our GSNE framework, which is highly  efficient and can work with heterogeneous types of nodes and features.  

\item Our comprehensive evaluation on a large real-estate dataset shows that for the house prediction task, combining geo-spatial embedded vectors learned by GSNE with the housing features results in consistently better prediction performance than raw feature only, regardless of the downstream regression model.

\end{itemize}
\section{Related Work}

In this section, we primarily discuss the existing works on house price predictions. Based on the working methodologies, we divide these works three categories: housing feature centric traditional approaches, machine learning based approaches, and location centric approaches, which are presented in Section~\ref{subsec:hp}. Later we discuss major existing works on networking embedding in Section~\ref{subsec:ne}

\subsection{House Price Predictions}
\label{subsec:hp}
\textbf{Housing Feature-Centric Traditional Approaches:} Most of the earlier price prediction models were based on Hedonic Regression~\cite{rosen1974hedonic}. Later this model has been extensively studied predicting prices of different areas and analyzing the effects of different factors\cite{trojanek2013measuring,yayar2014hedonic,krol2015application,ottensmann2008urban}. In Hedonic price model, houses are considered as aggregation of different attributes, where customers purchase this package of bundled attributes. Although it simplifies the prediction task, there are some notable shortcomings. It has been found that hedonic price coefficients of some attributes are not stable between locations, property types and age\cite{fletcher2000modelling}. Furthermore, issues like  model
specification procedures, independent variable interactions,
non-linearity and outlier data points inhibits price prediction performance in hedonic price models \cite{limsombunchai2004house}. 
Genetic algorithms have also been used in the study of house price prediction problem. In \cite{ng2008using}, the authors used a hybrid of genetic algorithm and SVM to predict house prices from different sets of features. In another work, Manganelli et. al.\cite{manganelli2015using} studied the potential of genetic algorithm in this problem domain. They also study the effect of geographical location from the viewpoint of genetic algorithm. In a recent work\cite{morano2018multicriteria}, Morano et. al use evolutionary polynomial regression to model house price prediction.

\textbf{Machine Learning (ML) Approaches:} Following the success of ML models in different prediction tasks, researchers started to employ different machine learning techniques for estimating housing prices. Wang et. al\cite{wang2008application} and Li et. al \cite{li2009svr} used SVM based regression for determinig the house price. Xin et. al used Lasso and Ridge Regression for predicting house prices\cite{xin2018modelling}. In \cite{limsombunchai2004house}, the author found that Artifical Neural Network(ANN) based model outperforms hedonic price models in out of sample predictions. Later, in another study\cite{ravikumar2017real}, the authors experimented with a wide varieties of ML techniques that include Artificial Neural Networks (ANN),
AdaBoost, Random forest, Gradient boosted trees, Multi Layer Perceptron, and Ensemble learning algorithms. They found that Gradient boosted trees yield the best performance in predicting house prices. Researchers also used modern deep learning based approaches to improve house price prediction performance. In \cite{piao2019housing}, the authors used convolutional neural network(CNN) for feature selection as well as price prediction. Feng et. al\cite{feng2015comparing} compared multi level modeling (MLM) approaches with ANN and found the MLM methods to be much superior compared to ANN. In another recent work~\cite{zhao2019deep}, the authors utilized property images alongside original tabular features, where they used CNN to extract features from those images and combined them with the transformed tabular features. They achieved an improved performance in price prediction by channeling this new set of features through XGBoost algorithm.

\textbf{Location-Centric Approaches:}
As researchers realized the impact of locations on house prices, several works focused on the spatial awareness of the prediction models in order to amplify the location effect. Standard Hedonic regression models assume the residuals to be independent of each other, yet it is found that those residuals show significant spatial dependency\cite{pace1998generalizing}. Over the years, several techniques have been proposed to introduce spatial awareness in the prediction models. In \cite{bourassa2003housing}, the authors experimented with a set of spatial submarkets defined by real estate appraisers. They compared different approaches of including neighbouring properties' residuals, separate submarket equations, etc. to take spatial dependence into account. On the other hand, Fletcher et. al. \cite{fletcher2000modelling} found that prediction from a model with postcode dummies perform slightly better than separate equations for each postcode. Geo-statical approaches have also been taken in some works \cite{dubin1998predicting,basu1998analysis}, where it was found to do well compared to ordinary least square (OLS) regression models. Bourassa et. al in \cite{bourassa2010predicting} found that geo-statistical model with
dis-aggregated submarket variables performed the best in predicting price while considering spatial dependency. On the other hand, Thibodeau et.al \cite{thibodeau2003marking} found that in neighbourhood level geo-statical models perform only slightly better compared to OLS model. Besides geo-statical approaches, lattice approaches have also been tried out. In \cite{bourassa2007spatial}, the authors found that in mass appraisal context, lattice models which include SAR (simultaneous autoregressive) and CAR (conditional autoregressive) models performed poorly compared to even simple OLS models which disregard spatial dependence. They concluded that including submarket variable in OLS context gives much better gain in accuracy compared to any other geo-statical or lattice methods. They also argued that it is much more practical given that hedonic models with submarket dummies are much easier to implement compared to geo-statical or lattice approaches while also giving better gains. In \cite{case2004modeling}, the authors experimented with OLS with location variables, geo-statical and several spatial statistics methods. They found that when accounted for the neighbouring residuals, all of those models showed similar results. Fik et. al\cite{fik2003modeling} on the other hand, showed that property characteristics as well as cartesian coordinates and submarket dummies were able to capture most of spatial dependence and gives significant improvement in prediction accuracy.

In another work,\cite{montero2018housing} Montero et. al considered parametric and semi-parametric spatial hedonic model variants to capture spatial auto-correlation. On the other hand, an unpublished technical report by Gao et. al\cite{gao2019location} partitioned their dataset for different task definitions based on different schemes such as distance to station, schools, etc and used multi-task learning approach for the partitioned dataset. A major limitation of these approaches is that it requires in depth domain analysis for choosing the submarket definition/partitioning scheme as effective partitioning scheme vary widely from dataset to dataset. Furthermore, to induce spatial lag in price prediction, traditional spatial statistics(e.g. Geostatical, SAR, CAR) approaches relied on heavy amount of feature engineering, which is required to be done by real estate appraisers. Recently, Jenkins et. al\cite{jenkins2019unsupervised} used multimodal data such as Satellite images, taxi mobility data, and categories of point of interests for generating embedding for different grid-partitioned region, which are later used to find overall price per sqft for houses in that region. This approach only gives an overview of the region and considers price per sqft. of houses, which is a coarse outline of price distribution. None of these methods could capture the intrinsic relationships between neighborhood POIs and the corresponding house, which is our key contribution of this paper.

\subsection{Network Embedding}
\label{subsec:ne}
To represent complex high dimensional network information in a low dimensional feature space, several techniques have been adopted by the researchers in recent times \cite{cai2018comprehensive}. Random walk based methods\cite{perozzi2014deepwalk,grover2016node2vec}, graph convolution\cite{kipf2016semi}, proximity objective centric LINE\cite{tang2015line} are among the notable ones. Recently, Gaussian Embedding has been shown to be quite effective in modeling the inherent uncertainty of real world data in \cite{bojchevski2017deep,hettige2020gaussian,zhu2018deep}. Since attributed geo-spatial network is complex and noisy in nature, we adopt Gaussian Embedding to handle the uncertainties effectively. However, the prior works in Gaussian Embedding only considers homogeneous or bipartite network, which is not straightforward to adapt for multipartite attributed geo-spatial networks.

\section{Methodology}

\begin{figure}[ht]
\begin{center}
\includegraphics[width=0.6\textwidth,height=0.6\textheight]{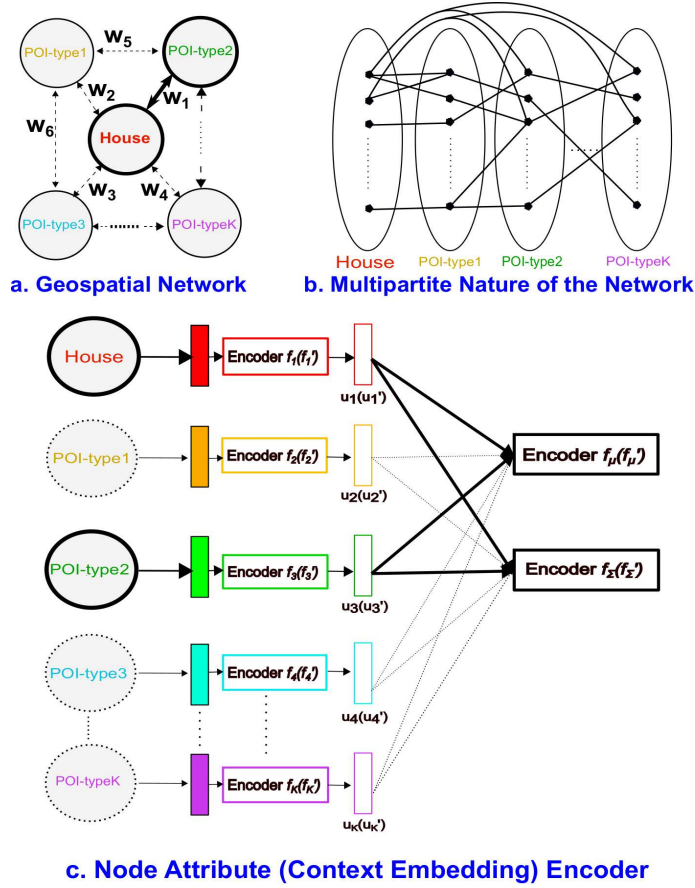}
\caption{ (a) An example geo-spatial network centred around a house.  (b) Illustration of the multipartite nature of the network, where each node partition represents either houses or a specific category of POI nodes (c) The overall architecture of GSNE, where the 
node attribute encoding is represented using $\{f_1, f_2, ..., f_K\}$, $\{u_1, u_2, ...,u_K\}$, and $\{f_{\mu}, f_{\sigma}\}$, and the second-order context encoding is represented as $\{f_{1}^{\prime}, f_{2}^{\prime}, ..., f_{K}^{\prime}\}$, $\{u_{1}^{\prime}, u_{2}^{\prime}, ...,u_{K}^{\prime}\}$, and $\{f_{\mu}^{\prime}, f_{\sigma}^{\prime}\}$.}

\label{fig:GSNE}
\end{center}
\end{figure}

In this section, we present our proposed geo-spatial network embedding (GSNE) approach. Since GSNE utilizes the notion of network embedding, given the housing data, GSNE converts the house and POI data into a geo-spatial network that is represented as an attributed, multipartite network. This attributed, multipartite network is then channeled through a neural network embedding pipeline, which embeds the nodes of the network in a Gaussian feature space, as inspired by previous work~\cite{bojchevski2017deep,hettige2020gaussian,zhu2018deep}. 
The high-level architecture of GSNE is depicted in Figure~\ref{fig:GSNE}, which will be discussed in subsequent sections.

Next, we discuss the problem formulation of geo-spatial network embedding in Section~\ref{subsec: network}. Then we present the details of heterogeneous node attribute encoding in Section~\ref{subsec:HNAE}. 
After that we present the structure embedding learning through both first- and second-order proximity (Section~\ref{subsec:structure}). 
We discuss our model optimization in Section~\ref{sec:optimization}, which will be followed by model training details for GSNE (Section~\ref{sec:Training}).

\subsection{Problem Definition}
\label{subsec: network}

Let $G = \{\mathcal{V}, \mathcal{X}, E, W\}$ represent our attributed, weighted, multipartite geo-spatial network, where, $\mathcal{V}$, $\mathcal{X}$, $E$, and $W$ represent nodes (vertices), node attribute matrices, edges, and edge weights, respectively. 
The set of nodes $\mathcal{V}$ comprises $K$ mutually exclusive subsets of nodes, i.e., $\mathcal{V} = V_1 \cup V_2 \cup \ldots \cup V_K$. 
The set of attribute matrices is represented as $\mathcal{X} = \{\boldsymbol{X}_1, \boldsymbol{X}_2,\ldots,\boldsymbol{X}_K\}$, such that each $\boldsymbol{X}_k\in\mathbb{R}^{D_k\times |V_k|}$ represents the attribute matrix for partition $k$, where $D_k$ is the dimension of the attributes and $|V_k|$ represents the number of nodes in partition $k$.
$W\in \mathbb{R}^{|E|}$ is a weight {vector} such that each edge $e_{ij}\in E$ between a pair of nodes $V_i$ and $V_j$ has a weight $w_{ij}\in W$. 
Note that without the loss of generality we use the terms `attributes' and `features' interchangeably.

\begin{table}
\caption{Basic notation}
\vspace{-6pt}
\centering
\begin{tabular}{c|p{6.1cm}}
\hline
{\bfseries Symbol} & {\bf Description}\\
\hline
%
$G$ & $G = \{\mathcal{V}, \mathcal{X}, E, W\}$ is our attributed, weighted, undirected geo-spatial network \\
$\mathcal{V}$ & The set of nodes $\mathcal{V} = V_1 \cup V_2 \cup \ldots \cup V_K$ having $K$ subsets of nodes \\
$\mathcal{X}$ & The set of attribute matrices $\mathcal{X} = \{\boldsymbol{X}_1, \boldsymbol{X}_2,\ldots,$ $\boldsymbol{X}_K\}$, each $\boldsymbol{X}_k \in \mathcal{X}$ represents the attribute matrix of node partition $V_k$\\
$E$ & Set of edges of geo-spatial network\\
$W$ & Weight vector of the edges  \\ 
$w_{ij}$ & $w_{ij} =\frac{1}{\delta(i,j)}$, $\delta(i,j)$ represents the distance between nodes $i,j \in \mathcal{V}$  \\
$L$ & Output embedding dimension \\
$\{\boldsymbol{W}_{k_1}, \boldsymbol{W}_{k_2}\}$ ($\{\boldsymbol{W}^\prime_{k_1}, \boldsymbol{W}^\prime_{k_2}\}$) & Weights of the attribute encoder of node partition $k$ for first(second) order proximity\\
$\{\boldsymbol{b}_{k_1}, \boldsymbol{b}_{k_2}\}$ ($\{\boldsymbol{b}^\prime_{k_1}, \boldsymbol{b}^\prime_{k_2}\}$) & Bias vectors of the attribute encoder of node partition $k$ for first(second) order proximity \\
$f_k(f^{\prime}_{k})$ & Attribute encoder for first(second) order proximity\\
$\boldsymbol{u_i}(\boldsymbol{u^\prime_i})$ & Intermediate representation of node $i$ after attribute encoding stage for first(second) order proximity\\
$\boldsymbol{W}_\mu, \boldsymbol{W}_\Sigma (\boldsymbol{W}^\prime_\mu, \boldsymbol{W}^\prime_\Sigma)$ & Mean and covariance encoder weights for first(second) order proximity \\
$\boldsymbol{b}_\mu, \boldsymbol{b}_\Sigma (\boldsymbol{b}^\prime_\mu, \boldsymbol{b}^\prime_\Sigma)$ & Mean and covariance encoder biases for first(second) order proximity \\

$f_\mu, f_\Sigma(f^{\prime}_{\mu}, f^{\prime}_{\Sigma})$ & Mean and covariance encoder for first(second) order proximity\\
$\boldsymbol{\mu_i}, \boldsymbol{\Sigma_i}$ $(\boldsymbol{\mu^\prime_i}, \boldsymbol{\Sigma^\prime_i})$ & Mean and covariance embedding representation of node $i$ for first order proximity(second order proximity)\\
$\boldsymbol{h_i}(\boldsymbol{h^\prime_i})$ & $\boldsymbol{h_i} = \mathcal{N}(\boldsymbol{\mu_i}, \boldsymbol{\Sigma_i})$ is the first order ($\boldsymbol{h^\prime_i} = \mathcal{N}(\boldsymbol{\mu^\prime_i}, \boldsymbol{\Sigma^\prime_i})$ is the second order) proximity embedding of node $i$\\

$D(\boldsymbol{h_i}, \boldsymbol{h_j})$ & Asymmetric KL-divergence between embeddings $\boldsymbol{h_i}$ and $\boldsymbol{h_j}$\\
$d(\boldsymbol{h_i}, \boldsymbol{h_j})$ & Symmetric KL-divergence between embeddings $\boldsymbol{h_i}$ and $\boldsymbol{h_j}$\\

\hline
\end{tabular}
\label{table:notation}
\vspace{-10pt}
\end{table}

We exploit the spatial information (latitude, longitude) of houses and POIs to generate the set of edges $E$. 
Intuitively, an edge is created between a pair of nodes only if they are located less than a certain distance threshold $\delta_{max}$. 
For each edge $e_{ij} \in E$, its weight $w_{ij}$ is calculated as $w_{ij} =\frac{1}{\delta(i,j)}$, where $\delta(i,j)$ represents the distance between the two nodes. 
Moreover, we design our network to be undirected, thus $e_{ij} = e_{ji}$ and $w_{ij} = w_{ji}$. 

For our geo-spatial network, the partition $V_1$ denotes houses, and the other partitions $V_2,\ldots,V_K$ denote different types of POIs such as regions, schools and train stations. Figure~\ref{fig:GSNE} (a) and (b) show an example of geo-spatial network and its multipartite representation, respectively.

The GSNE model learns low-dimensional Gaussian embeddings $\boldsymbol{h}_i$ for each node $v_i\in\mathcal{V}$, i.e.\ $\boldsymbol{h}_i = \mathcal{N}(\boldsymbol{\mu}_i,\boldsymbol{\Sigma}_i)$, where  $\boldsymbol{\mu}_i\in \mathbb{R}^L$, $\boldsymbol{\Sigma}_i\in\mathbb{R}^{L\times L}$. 
$L$ is the embedding dimension and $L\ll|\mathcal{V}|$ and $L\ll D_k$ for each $k\in1,\ldots,K$. 
Intuitively, nodes that are similar to each other and are close in the original multipartite network are also close in the embedding space.

\subsection{Heterogeneous Node Attribute Encoding}
\label{subsec:HNAE}
We employ neural network based encoders to learn the embeddings of node features of the $K$ partitions 
in the network $G$.
A straightforward approach for encoding node features in a multipartite network is to concatenate feature vectors of different partitions with zero padding to create a \emph{global} feature vector, which is then fed into a single encoder to produce the initial node embeddings.
This approach has been shown to work well for bipartite networks (where $K=2$) in prior work~\cite{hettige2020gaussian}. 
However, this strategy may face issues in a complex multipartite network like ours, as the dimension of the combined feature vector becomes exceedingly large and sparse that it induces suboptimal model convergence as found in our experiments. 

To tackle this challenge, GSNE uses separate encoders to handle different partitions of nodes{, which is followed by a global Gaussian encoder that projects distinct types of POI nodes in the same Gaussian embedding space}. By doing so, we solve the sparsity issue while projecting the final embedding into the same Gaussian embedding space.
Specifically, we employ $K$ different encoders $\{f_k\in \boldsymbol{X}_k\rightarrow \mathbb{R}^{l \times |V_k|}\}_{k \in \{1,2,\ldots,K\}}$, where each encoder $f_k$ projects attributes $\boldsymbol{X}_k$ of nodes in partition $V_k$ to $l$-dimensional embedding space $\mathbb{R}^{l \times |V_k|}$. 
Each encoder $f_k$ generates encoded representations $\boldsymbol{u}_i$ for a feature vector $\boldsymbol{x_i} \in \boldsymbol{X}_k$ using rectified linear units (ReLU) as follows:

\begin{align}
    \boldsymbol{u}_i &= f_k(\boldsymbol{x_i}) = ReLU(\boldsymbol{W}_{k_2}\boldsymbol{z}_i + \boldsymbol{b}_{k_2}) \text{, where} \\
    \boldsymbol{z}_i &= ReLU(\boldsymbol{W}_{k_1}\boldsymbol{x_i} + \boldsymbol{b}_{k_1}) \nonumber
\end{align}
where $\boldsymbol{W}_k = \{\boldsymbol{W}_{k_1}, \boldsymbol{W}_{k_2}\}$ are the weight matrices and $\boldsymbol{b}_k = \{\boldsymbol{b}_{k_1}, \boldsymbol{b}_{k_2}\}$ are the bias vectors for partition $V_k \in \mathcal{V}$. Here, $\boldsymbol{W}_{k_1} \in \mathbb{R}^{l^{\prime} \times D_k} $, $\boldsymbol{W}_{k_2} \in \mathbb{R}^{l \times l^{\prime}} $, $\boldsymbol{b}_{k_1} \in \mathbb{R}^{l^{\prime}}$, and $\boldsymbol{b}_{k_2} \in \mathbb{R}^{l}$.

The attribute embeddings $\boldsymbol{u}_i$ are then channeled through a Gaussian encoder to obtain the final Gaussian embedding $\mathcal{N}(\boldsymbol{\mu}_i, \boldsymbol{\Sigma}_i)$. 
Although we use separate encoders for each node type to handle different partitions, we employ a common global Gaussian encoder $(f_\mu, f_\Sigma)$ for learning the final representations of all nodes. As a result, the common Gaussian encoder allows us project all nodes in the same $L$-dimensional Gaussian space. 
Since we are projecting nodes of different partitions in the same Gaussian space, we are essentially allowing the nodes to know their own neighbourhood while learning from the global network structure.
Here we generate Gaussian embedding $\boldsymbol{h}_i = (\boldsymbol{\mu}_i, \boldsymbol{\Sigma}_i)$ from the intermediate representations as follows:

\begin{align}
    \label{eq:02}
    \boldsymbol{\mu}_i &= f_\mu(\boldsymbol{u}_i) = ReLU(\boldsymbol{W}_{\mu}\boldsymbol{u}_i + \boldsymbol{b}_{\mu})  \\
    \label{eq:03}
    \boldsymbol{\Sigma}_i &= f_\Sigma(\boldsymbol{u}_i) = ELU(\boldsymbol{W}_{\Sigma}\boldsymbol{u}_i + \boldsymbol{b}_{\Sigma}) + 1
\end{align}
where $\boldsymbol{W}_\mu,\boldsymbol{W}_{\Sigma} \in \mathbb{R}^{L\times l}$ and $\boldsymbol{b}_\mu,\boldsymbol{b}_{\Sigma} \in \mathbb{R}^{L}$, and the exponential linear unit (ELU)~\cite{clevert2015fast} is the activation function in the covariance encoder.

\subsection{Network Structure Embedding}\label{subsec:structure}
GSNE learns from the structure of the multipartite network by considering both first-order and second-order proximity between each pair of connected partitions. 
First-order proximity learns from the direct connections of a pair of nodes across partitions, while second-order proximity learns from nodes that are connected through an intermediate node. The aim of this learning is to capture local neighbourhood context as well as global connectivity in the whole network.
In the following subsections we describe how such structural information is captured in GSNE.

\subsubsection{First-order Proximity in Geospatial Network}
As indicated in previous work~\cite{tang2015line,wang2016structural}, the \emph{first-order proximity} represents the local pairwise proximity between two nodes. 
In other words, if an edge connects a pair of nodes $(i,j)$, they have a positive first-order proximity. 
While in any network it is intuitive to generate similar embeddings for two nodes with positive first-order proximity, in the geospatial domain it has additional significance. 
Consider the following scenario. 
If a house $i$ has an edge with a nearby school $j$, the first-order proximity essentially tries to generate similar embeddings for them. 
If this same house $i$ also has an edge with a train station $k$ in its neighbourhood, it also tries to keep them close in the embedding space. 
The outcome is that, the first-order proximity lets the model learn from the geospatial connectivity of a node's \emph{local} neighbourhood amenities. 
Thus, the model essentially learns to embed the neighbourhood context of a house, which customers may likely take into consideration while making purchase decisions of a house and thus influence the price of the house. 

\subsubsection{Second-order Proximity in Geospatial Network}
While the first-order proximity can effectively capture the local neighbourhood context of a node, it fails to capture latent similarity of two nodes beyond their immediate neighbourhood when there are no direct edges between them. 
However, consider the following scenario. If two houses are located in two geographically distant locations, yet both of them are connected to the same highly-rated schools and transportation facilities (e.g.\ train stations) nearby, their geo-spatial embedding should also be similar. 
The \emph{second-order proximity} between two nodes in a network essentially capture  the similarity between their neighbourhood network structure. 
Thus, to take advantage of this fact, GSNE uses second-order proximity so that geographically distant houses having similar neighbourhood structure are located closer in the embedding space. 
In other words, it helps the model learn the \textbf{global} neighbourhood connectivity of the network. 

\subsubsection{Proximity Objectives}
We adopt the widely used strategy of LINE~\cite{tang2015line} to compute our first- and second-order proximities. As we embed nodes as Gaussian distributions, we employ KL-divergence as our dissimilarity measure. 
Since our network is undirected and KL-divergence is asymmetric in nature, we consider both directions of the edges. 
i.e.\ we compute $D(\boldsymbol{h}_i$, $\boldsymbol{h}_j$) + D($\boldsymbol{h}_j$, $\boldsymbol{h}_i$) the KL-divergence from both directions as suggested in previous work~\cite{bojchevski2017deep}.
Let $\boldsymbol{h}_i$ and $\boldsymbol{h}_j$ represent the Gaussian representations of nodes $i$ and $j$, and 
{$d(h_i, h_j) = D(\boldsymbol{h}_i$, $\boldsymbol{h}_j$) + D($\boldsymbol{h}_j$, $\boldsymbol{h}_i$)} represent their KL-divergence dissimilarity measure.

\textbf{First-Order Proximity:} For each $(i,j) \in E$, we take the joint probability between $i$ and $j$ as:
\begin{equation}
    \label{eq:p1}
    p_1(i, j) = \frac{1}{1 + \exp{(d(\boldsymbol{h}_i, \boldsymbol{h}_j))}}
\end{equation}

With this joint probability, we take the first-order proximity objective $O_1$ as follows.

\begin{equation}
    \label{eq:05}
    O_1 = - \sum_{(i,j) \in \emph{E}} w_{ij} \log p_1(i,j)
\end{equation}
where $w_{ij}$ is the edge weight.

\textbf{Second-order Proximity:}  For second-order proximity, each node $i \in \mathcal{V}$ requires both attribute embedding $\boldsymbol{h}_i$ and context embedding $\boldsymbol{h}^{\prime}_i$, which is treated as a context of other nodes.
However, the traditional definition of second-order proximity~\cite{tang2015line} is defined only on homogeneous and bipartite networks, but not multipartite networks like ours. 
This causes a problem in negative node sampling (discussed in Section \ref{sec:optimization}) which inhibits the model from convergence. 
Thus, we modify the second-order proximity as follows. 

For each directed edge $(i,j)\in E$ (undirected edges can be treated as two edges in opposite directions), where $i \in V_p$ and $j \in V_q$ are nodes from two partitions $V_p$ and $V_q$ respectively, the probability of the context of node $j$ generated by node $i$ is:

\begin{equation}
\label{eq:06}
p_2(j | i) = \frac{\exp{(-d(\boldsymbol{h}_i, \boldsymbol{h}^\prime_{j}))}}{\Sigma_{\hat{i} \in V_q} \exp{(-d(\boldsymbol{h}_i, \boldsymbol{h}^\prime_{\hat{i}}))}}
\end{equation}
where $\hat{i}$ range over all nodes in partition $V_q$. 

With this new definition of $p_2$, we define our second-order proximity objective as in LINE~\cite{tang2015line}:
\begin{equation}
    \label{eq:07}
    O_2 = - \sum_{(i,j) \in E} w_{ij} \log p_2(j | i)
\end{equation}

In order to generate the context embedding $\boldsymbol{h}^\prime_i = (\boldsymbol{\mu}^\prime_i, \boldsymbol{\Sigma}^\prime_i)$, we need a separate set of encoders. 
We define $f^\prime_k=(\boldsymbol{W}^\prime_k, \boldsymbol{b}^\prime_k)$ where $\boldsymbol{W}^\prime_k = \{\boldsymbol{W}^\prime_{k_1}, \boldsymbol{W}^\prime_{k_2}\}$ and $\boldsymbol{b}^\prime_k = \{\boldsymbol{b}^\prime_{k_1}, \boldsymbol{b}^\prime_{k_2}\}$ are the weight matrices and bias vectors for partition $V_k \in \mathcal{V}$. For Gaussian context embedding, with model parameters $f^\prime_\mu = (\boldsymbol{W}^\prime_\mu, \boldsymbol{b}^\prime_\mu)$, $f^\prime_\Sigma = (\boldsymbol{W}^\prime_\Sigma, \boldsymbol{b}^\prime_\Sigma)$, we generate context embedding as following:

\begin{align}
    \label{eq:08}
    \boldsymbol{u}^\prime_i &= f^\prime_k(\boldsymbol{x}_i) = ReLU(\boldsymbol{W}^\prime_{k_2}\boldsymbol{z}^\prime_i + \boldsymbol{b}^\prime_{k_2})  \text{ where}\\
    &\quad \boldsymbol{z}^\prime_i = ReLU(\boldsymbol{W}^\prime_{k_1}\boldsymbol{x}_i + \boldsymbol{b}^\prime_{k_1}) \nonumber\\
    \label{eq:09}
    \boldsymbol{\mu}^\prime_i &= f^\prime_\mu(\boldsymbol{u}^\prime_i) = ReLU(\boldsymbol{W}^\prime_{\mu}\boldsymbol{u}^\prime_i + \boldsymbol{b}^\prime_{\mu})  \\
    \label{eq:10}
    \boldsymbol{\Sigma}^\prime_i &= f^\prime_\Sigma(\boldsymbol{u}^\prime_i) = ELU(\boldsymbol{W}^\prime_\Sigma \boldsymbol{u}^\prime_i + \boldsymbol{b}^\prime_{\Sigma}) + 1
\end{align}

Similar to the encoders in Section \ref{subsec:HNAE}, here $\boldsymbol{W}^\prime_{k_1} \in \mathbb{R}^{l^{\prime} \times D_k} $, $\boldsymbol{W}^\prime_{k_2} \in \mathbb{R}^{l \times l^{\prime}} $, $\boldsymbol{b}^\prime_{k_1} \in \mathbb{R}^{l^{\prime}}$, $\boldsymbol{b}^\prime_{k_2} \in \mathbb{R}^{l}$, $\boldsymbol{W}^\prime_\mu,\boldsymbol{W}^\prime_{\Sigma} \in \mathbb{R}^{L \times l}$ and $\boldsymbol{b}^\prime_\mu,\boldsymbol{b}^\prime_{\Sigma} \in \mathbb{R}^{L}$

\subsection{Model Optimization}
\label{sec:optimization}
Since the objective function in Eq.~\ref{eq:07} requires the summation over entire set of nodes of the same type while calculating $p_2$, it is computationally prohibitive. 
To alleviate this issue, we use negative sampling~\cite{tang2015line}. With this negative sampling technique employed, objective $O_2$ becomes:

\begin{align}
    \label{eq:11}
\begin{split}
    O_2 = &\sum_{(i,j) \in \emph{E}, i \in V_p, j \in V_q} \big( \log\;{\sigma(-d(\boldsymbol{h}_i, \boldsymbol{h}_j^\prime))}\ +\\
    &\sum_{n = 1}^{N} \mathbb{E}_{v_n \thicksim P_{qn}(v)} \log \;{\sigma(d(\boldsymbol{h}_i, \boldsymbol{h}^\prime_{v_n}))} \big)
    \end{split}
\end{align}

Here the first term optimizes the positive edges whereas the second term is concerned with negative edges drawn from a noise distribution $P_{qn}(v) \propto d_v^{\frac{3}{4}}$, where $v \in
V_q$ and $d_v$ is the degree of the node(since the network is undirected). 
We use this negative sampling strategy also in the calculation of $O_1$ in Eq.~\ref{eq:05}, which is similar to Eq.\ref{eq:11} except $\boldsymbol{h}^\prime$ in the equation will be changed to $\boldsymbol{h}$. 

When $O_2 $ is optimized, the gradient gets multiplied by the edge weight. Since the edge weights are set to be the inverse of geographical distance, edges in our geo-spatial network may have high weight variance. This may induce exploding gradient during training phase. A naive solution to this problem is unwrapping an edge of weight $w$ units into $w$-binary unweighted edges so that the whole graph can be regarded as unweighted. Yet, it is very inefficient from the memory perspective of the network.

To avoid exploding gradient without compromising memory efficiency, we use edge sampling from an alias table as in \cite{tang2015line}, where we efficiently sample positive edges according to the distribution of the weights of the edges using alias table.

\begin{figure}
\centering
\includegraphics[width=\textwidth,height=0.25\textheight]{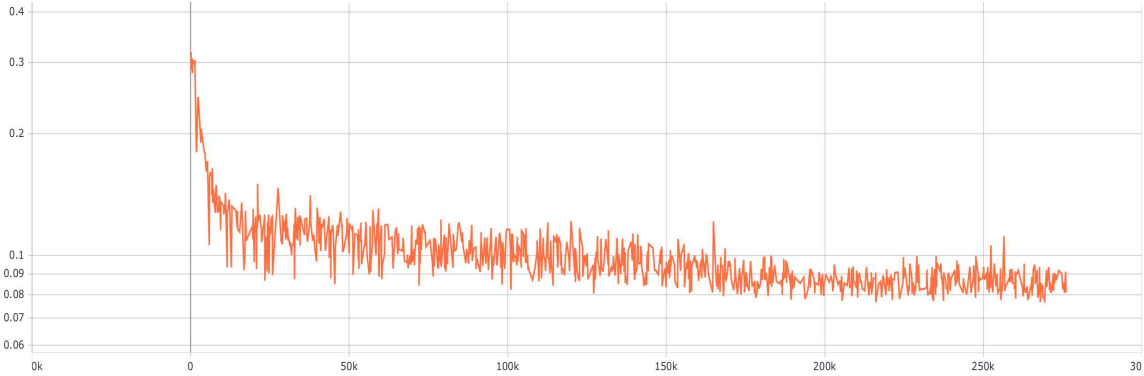}
\centering
\caption[]{Training Loss Curve for GSNE.}
\label{fig:loss_curve}

\end{figure}

\subsection{Model Training}
\label{sec:Training}
The training algorithm of GSNE is presented in Algorithm \ref{algo:GSNE}. 
Here we first initialize the parameters of our network described previously. 
Then we select an edge set $E_i \subseteq E$ where $E_i$ represents connections between nodes between two partitions. Subsequently, we sample a batch of positive edges from $E_i$ as in LINE~\cite{tang2015line}. 
For each of the positive edges, we also sample $N$ negative edges. 
Later we calculate the Gaussian embeddings of the node pairs of sampled edges and calculate the batch loss through back propagation and updating the parameters $\theta$.

\begin{algorithm}[htb]
\DontPrintSemicolon
  
  \KwIn{Network $G = \{\mathcal{V}, \mathcal{X}, E, W\}$, proximity, number of negative samples $N$, batch size $b$, total iterations $T$ }
  \KwOut{Embedding $\boldsymbol{h}_i$ for each node $i\in \mathcal{V}$ }
  Let $\theta = \{f_{i}\}_{i \in \{1,2,\ldots,K\}} \bigcup \{f_\mu, f_\Sigma\}$

  \If{proximity = second-order}
    {
        $\theta = \theta \bigcup \{{f^\prime_{i}}\}_{i\in\{1,2,\ldots,K\}} \bigcup \{f^\prime_\mu, f^\prime_\Sigma\}$
    }
    
    Initialize $\theta$\\
    \For{iterations = $0 \to T$}
    {
    \For{\textbf{each} edge 
    set $E_i \subseteq  E$}
    {
    $batch = FetchBatch(E_i, b)$\\
    \For{\textbf{each} edge $(i,j) \in batch$}{
    Calculate $\boldsymbol{h}_i$ (Eq. \ref{eq:02}, \ref{eq:03})\\
    $Neg(i) = NegativeSampling((i,j), N, E_i)$\\
    \If{proximity = first-order}{
    Calculate $\boldsymbol{h}_j, \{\boldsymbol{h}_{\nu_k}\}_{\nu_k \in Neg(i)}$ \hfill$\triangleright$ (Eq. \ref{eq:02}, \ref{eq:03})\\
    } 
    \Else{
    Calculate $\boldsymbol{h}^\prime_j, \{\boldsymbol{h}^\prime_{\nu_k}\}_{\nu_k \in Neg(i)} $ \hfill$\triangleright$ (Eq. \ref{eq:09}, \ref{eq:10})\\
    }
    }
    }
    \If{proximity = first-order}{
    Calculate $O_1$ \hfill $\triangleright$ (Eq. \ref{eq:05})
    } 
    \Else{
    Calculate $O_2$ \hfill $\triangleright$ (Eq. \ref{eq:11})
    }
    Update the parameters $\theta$ in the backpropagation stage
    }
\caption{The GSNE algorithm}
\label{algo:GSNE}
\end{algorithm}

Selection of a bipartite edge set $E_i \subseteq E$ in each iteration is an important factor to consider here as it may impact model performance. We experimented with three strategies: (1) randomly selecting pairs of connected partitions, (2) iteratively alternating between pairs of connected partitions in every iteration, and (3) selecting a pair of connected partitions per 100 iterations and alternate. 
We found that strategy (2), iterative alternation between bipartite edge sets, is the strategy yielding the best result overall. 
Even though iterative alternation causes the loss to be jumpy, the overall trend of the loss decreases as shown in Figure~\ref{fig:loss_curve}. 
On the other hand, the other two strategies lead towards early convergence to local optima, resulting in higher training loss.

\section{Experiments}

To evaluate the efficacy of our Geo-spatial Network Embedding (GSNE) method\footnote{https://github.com/sarathismg/gsne}, we apply it with a number of state-of-the-art regression models for the house price prediction task on a large real-estate and POI datasets of Melbourne, Australia. Specifically, our GSNE model is trained to obtain embeddings of houses. Then, we train each of the house price prediction (regression) models by concatenating our generated GSNE embeddings with the raw housing features (referred to as Raw + GSNE). We compare the prediction performance of our method (Raw + GSNE) against the same regression models trained on raw housing features only as described in Section~\ref{sec:dataset} (referred to as ``Raw"). To ensure the baseline, i.e.,  ``Raw" method contains spatial lag in the modeling, location details are also included along with the core housing features.

As the downstream regression models, we have chosen some of the best regression models for house price prediction competition in Kaggle~\cite{noauthor_stacked_nodate}, the recent house prediction models in \cite{xin2018modelling,xiong2019improve,ravikumar2017real}, and well known regression models such as LightGBM~\cite{ke2017lightgbm}, XGBoost~\cite{chen2016xgboost} and Gradient Boosting~\cite{ravikumar2017real}.

In the following, we first present the details of the dataset and the generation of geo-spatial network in Section~\ref{sec:dataset}. 
We then discuss our performance metrics for evaluating different algorithms in Section~\ref{subsec:pm}, followed by a discussion on the experimental setup in Section~\ref{sub:setup}. 
We analyze our house price prediction results in Section~\ref{subsec:pp}, including an ablation study on the effect of various components in our method. 
Finally, a qualitative analysis of the embeddings are presented using visualizations in Section~\ref{subsection:vis}. 

\subsection{Dataset Description}
\label{sec:dataset}
We conducted our experiments on the house transaction records obtained from a real-estate web site\footnote{https://www.realestate.com.au/} for Melbourne, which is the second largest city in Australia by population. We extracted a total of the 52,851 house transaction records of years from 2013 to 2015. 
Our dataset also includes the three types of POIs: \textit{regions}, \textit{schools}, and \textit{train stations} and their corresponding features. 
Houses are situated in regions which capture the geographical contextual information about houses.
Intuitively, information about nearby schools and train stations may influence house prices. 
Our dataset contains information of the 13,340 regions, 709 schools, and 218 train stations. 

\subsubsection{House and POI Features}

\textbf{Housing Features:} Our dataset contains information about a wide range of housing features. In total, we consider \textbf{43} housing features for each house for in depth exploration of the effect of GSNE. To the best of our knowledge, none of the prior works considered such a wide range of feature sets in a large dataset like ours for house price prediction task. Although the dataset in Kaggle competition \cite{noauthor_stacked_nodate} has 86 features, it has only 3000 samples in total and lots of columns are highly sparse rendering only a few of those columns truly useful. Besides, no information regarding neighbourhood amenities is available in that dataset. In our dataset, each house record contains information ranging from basic housing features like number of bedrooms, number of bathrooms, number of parking spaces, location, type of property, etc. to detailed facility features like air-conditioning, balcony, city-view, river-view, swimming, tennis-court, etc. These features are listed in detail in Table \ref{table:housing}.

\begin{table}

\caption{Housing Features}
\vspace{-6pt}
\centering
\begin{tabular}{c|p{6.1cm}}
\hline
%
Number of bedrooms & Fireplace  \\
Number of bathrooms & Fully fenced\\
Parking & Gas heating\\
Property type & Gym\\
Transaction date & Heating\\
Agency & Intercom\\
Latitude & Laundry\\
Longitude & Mountain\\
Air Conditioning & Park\\
Alarm & Swimming pool\\
Balconey & Renovated\\
BBQ & River view\\
City view & Rumpus room\\
Adjacency to schools & Sauna\\
Adjacency to shops & Study rooms\\
Adjacency to transport & Sun room\\
Courtyard & System heating\\
Number of dining rooms & Tennis court\\
Dish wash & Water views\\
Ducted & Wordrobe\\
Ensuite & Total additional features \\
Family rooms & \\

\hline
\end{tabular}
\label{table:housing}
\vspace{-10pt}
\end{table}

\noindent\textbf{Region Features:} Our dataset contains Melbourne region information at SA1 level\footnote{https://www.abs.gov.au/}. SA1 is the most granular unit for the release of census data of Australia~\cite{noauthor_what_2018}. The SA1 data typically has a population of 200 to 800 people with an average of 400 people per region. For each region, our dataset contains comprehensive information about the number of residents, average age, median personal income, percentage of Australian citizens, educational qualification, median house rent, location as the centroid of the region, etc. Since these aspects can be useful for determining house prices, we consider all of them as the features for regions.

\noindent \textbf{School Features:} The schools in our dataset are attributed with the type of school (primary or secondary), school category by gender(single gender or co-ed), ranking, location, number of students, zone restrictions, number of students enrolled in Victorian Certificate of Education(VCE), percentage of students securing 40\% marks, etc.\footnote{https://bettereducation.com.au/}

\noindent\textbf{Train Stations:} The train stations in the dataset contain information about their location and average time to reach to other stations.\footnote{http://developers.google.com/maps/}

\subsubsection{Dataset Pre-processing}
\label{subsec:preprocess}

\begin{figure}[ht]
\begin{center}
\includegraphics[width=\textwidth]{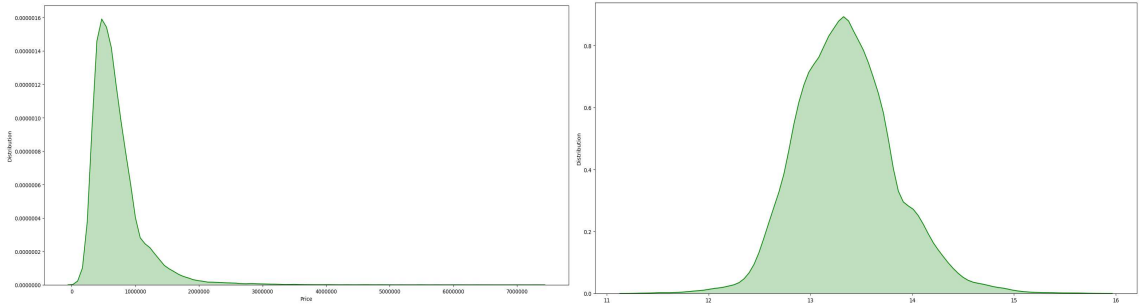}
\caption{Price distribution of the houses before and after log-normalization.}
\label{fig:preproc}
\end{center}
\end{figure}

From Figure \ref{fig:preproc}, we can see that the price distribution is skewed to the right. Since these skewed data  may induce higher influence on the error calculation of the more expensive houses, we apply the widely-used log-normalization to the prices as done in previous works \cite{gao2019location,xiong2019improve}. All performance measures are calculated on these normalized prices. 
 
For \textit{train stations} and \textit{schools}, we filled the missing values with the mean of the corresponding feature, since there are only 709 schools and 218 train stations. Categorical variables have been handled by one-hot encoding. For feature standardization, we use zero-mean, unit-variance on our dataset.

\subsubsection{Geo-spatial Network Generation}
From the dataset, we build a geo-spatial network using houses and the three different types of POIs: regions, schools, and train stations. We generate the network by considering the following edges: \textit{House-Region, House-School, House-Train Station, and School-Train Station}. 
In House-Region edges, every house in a region will be connected to the corresponding region. For House-School edges, we connect a house to all schools which are located in one kilometer radius. If there is no school found within a kilometer range we connect the house to the nearest school. Similarly, we form the edges between Houses and Train Stations. Apart from these three types of house-POI connections, we also include \textit{School-Train Station} edges as these can help to model the transport options to a school from a house via train. In addition, we also  maintain \textit{Train Station-Train Station} edges to keep the whole geo-spatial network connected. Weights for all types of edges are based on the Euclidean distance between the two nodes of the network. Since our primary focus is to create geo-spatial embeddings for house price prediction, we disregard the edges among POIs that seemingly do not have much impact in our problem domain. 

\subsection{Performance Metrics}
\label{subsec:pm}
Following the prior works in this area, we use `mean absolute error' (MAE) and `root mean squared error' (RMSE) as our metrics to evaluate our embedding model. 
They are defined as:
\DeclarePairedDelimiter\abs{\lvert}{\rvert}
\DeclarePairedDelimiter\norm{\lVert}{\rVert}

\begin{align}
	MAE &= \frac{1}{N}\sum_{i=1}^{N}|z_{i} - \hat{z}_i|\\
    RMSE &= \sqrt{\frac{1}{N}\sum_{i=1}^{N}\left( z_{i} - \hat{z}_i\right)^{2}}
\end{align}

Here $z_i$ represents the ground truth price and $\hat{z}_i$ represents the predicted house price. $N$ represents the number of samples. However, these metrics introduce a problem as higher prices influence the metrics much higher than the others\cite{xiong2019improve}. To mitigate this issue, we use the \emph{logarithmic} form of \textit{sales (sold) price} in prediction as described in Section \ref{subsec:preprocess}.

\subsection{Experimental Setup and Model Building}
\label{sub:setup}
All experiments were performed on a server with 16GB memory, and a 12GB NVIDIA Tesla P100 GPU. 

The dimension of our Gaussian embedding is set as $L = 32$. 
The total number of iterations is set as $T = 3,00,000$. 
The batch size is set as $b = 128$, and the number of negative samples is set as $N = 5$. 
The above sets of parameters are chosen based on the empirical evaluation with our dataset. 
We divided our dataset into two parts by using stratified random sampling, into 80\% for unsupervised training of the embedding model GSNE and 20\% for testing the embedding model. 
Since the test set is completely unseen in the training phase of the model, a good performance on the test set usually indicates that the embedding model successfully generalizes to unseen nodes, i.e., the model is \textit{inductive}. 

\subsection{House Price Prediction Results}
\label{subsec:pp}

\begin{table}[h]
\begin{center}
\caption{\label{tab:Price_Table}Housing Price Prediction Performance Comparison. Best performed is bolded for each model. }
\vspace{-2mm}
\resizebox{\linewidth}{!}{%
\begin{tabular}{lccccccccc}
    \toprule
     {\multirow{2}{*}{Metric}} & {\multirow{2}{*}{Method}} & \multirow{2}{*}{Lasso} & Elastic & Kernel & Gradient & XGBoost & \multirow{2}{*}{LGBM} & Avg.(KRR, & Meta-model\\
     & & & Net & Ridge & Boosting & (XGB) & & GBoost, XGB) & Stacking\cite{xiong2019improve}\\
    \midrule
    
    \multirow{7}{*}{MAE} & Raw & 0.251 & 0.245 & 0.149 & 0.136 & 0.143 & 0.135 & 0.140 & 0.135\\
    & Raw+ & \multirow{2}{*}{0.220} & \multirow{2}{*}{0.216} & \multirow{2}{*}{0.141} & \multirow{2}{*}{0.128} & \multirow{2}{*}{0.133} & \multirow{2}{*}{{0.129}} & \multirow{2}{*}{0.130} & \multirow{2}{*}{{0.129}}\\
    & GSNE($1^{st}$) &  &  &  &  & & & & \\
    & Raw+ & \multirow{2}{*}{0.247} & \multirow{2}{*}{0.241} & \multirow{2}{*}{0.137} & \multirow{2}{*}{0.126} & \multirow{2}{*}{0.133} & \multirow{2}{*}{\textbf{0.127}} & \multirow{2}{*}{0.130} & \multirow{2}{*}{0.129}\\
    & GSNE($2^{nd}$) &  &  &  &  & & & & \\
    & Raw+GSNE & \multirow{2}{*}{\textbf{0.209}} & \multirow{2}{*}{\textbf{0.205}} & \multirow{2}{*}{\textbf{0.135}} & \multirow{2}{*}{\textbf{0.125}} & \multirow{2}{*}{\textbf{0.132}} & \multirow{2}{*}{\textbf{0.127}} & \multirow{2}{*}{\textbf{0.128}} & \multirow{2}{*}{\textbf{0.128}}\\
    & ($1^{st}+2^{nd}$) &  &  &  &  & & & & \\
    \midrule
    
    \multirow{7}{*}{RMSE} & Raw & 0.333 & 0.331 & 0.206 & 0.195 & 0.200 & 0.190 & 0.197 & 0.190\\
    & Raw+ & \multirow{2}{*}{0.295} & \multirow{2}{*}{0.291} & \multirow{2}{*}{0.196} & \multirow{2}{*}{0.184} & \multirow{2}{*}{0.188} & \multirow{2}{*}{0.182} & \multirow{2}{*}{0.185} & \multirow{2}{*}{{0.185}}\\
    & GSNE($1^{st}$) &  &  &  &  & & & & \\
    & Raw+ & \multirow{2}{*}{0.339} & \multirow{2}{*}{0.334} & \multirow{2}{*}{0.191} & \multirow{2}{*}{0.182} & \multirow{2}{*}{0.188} & \multirow{2}{*}{\textbf{0.180}} & \multirow{2}{*}{0.184} & \multirow{2}{*}{0.184}\\
    & GSNE($2^{nd}$) &  &  &  &  & & & & \\
    & Raw+GSNE & \multirow{2}{*}{\textbf{0.290}} & \multirow{2}{*}{\textbf{0.289}} & \multirow{2}{*}{\textbf{0.190}} & \multirow{2}{*}{\textbf{0.181}} & \multirow{2}{*}{\textbf{0.187}} & \multirow{2}{*}{\textbf{0.180}} & \multirow{2}{*}{\textbf{0.182}} & \multirow{2}{*}{\textbf{0.183}}\\
    & ($1^{st}+2^{nd}$) &  &  &  &  & & & & \\
    \midrule
    
\end{tabular}}
\end{center}
\end{table}

We evaluate the effectiveness of our embeddings by using them as features to train various regression models. 
Specifically, we compare the performance of each regression model trained with two sets of features: with the house feature only (referred to as ``Raw'') and with the concatenation of the Raw features and our embeddings (referred to as ``Raw+GSNE''). We also consider three variants of the embeddings: first-order proximity only ($1^{st}$), second-order proximity only ($2^{nd}$), and both the first- and second-order ($1^{st} + 2^{nd}$) proximities. 
In each case, we concatenate these embeddings with the original raw features and use them to train the downstream regression models.

To ensure our comparison baseline ``Raw'' also include sufficient spatial lag in it's price model, we add the location details for the houses. We take this as representative of spatially aware prediction model as in \cite{fik2003modeling} since modern learning algorithms can effectively capture spatial dependence without any complex feature engineering required in OLS hedonic price models. We also experimented by adding postcode dummies as in  \cite{fletcher2000modelling}, although it did not provide any improvement in performance over location details. Hence, we did not include postcode features in ``Raw" feature set to ensure optimal performance in our comparison baseline while accurately capturing spatial dependence.

We consider a wide range of learning models to compare the performance. 
We use a number of widely-used regression models, including Lasso and Ridge regression as in~\cite{xin2018modelling}, and Random Forest Regression, Elastic Net Regression, and Kernel-Ridge Regression. 
Furthermore, we also train a number of state-of-the-art models including Gradient Boosting, XGBoost~\cite{chen2016xgboost}, LightGBM~\cite{ke2017lightgbm}, which have recently been shown to yield good performance in this problem. 
Finally, we also evaluate the ensembles of these models, including Averaging and Stacking with meta-model\cite{xiong2019improve}. 
For Stacking, we used Gradient Boosting, XGBoost, and Kernel-Ridge Regression as first stage models, and Kernel-Ridge Regression as our meta model. 

\subsubsection{Result Summary}

Table \ref{tab:Price_Table} summarizes our house prediction results, where we observe that for all models, the Raw+GSNE-based house prediction results consistently outperform that with the Raw features(including location information) only. We also observe that the less expressive variants of our embeddings, first-order ($1^{st}$) or second-order($2^{nd}$) proximity, also outperform Raw in all cases. 

From Table \ref{tab:Price_Table}, we observe that with MAE, our best performing embeddings, i.e., Raw+GSNE\newline ($1^{st}$+$2^{nd}$), outperform Raw with different downstream models by a notable margin ranging from 5.2\% to 16.73\%. 
Moreover in the best performing model, Gradient Boosting, Raw+GSNE outperforms Raw by 8.1\%. 
For RMSE, Raw+GSNE($1^{st}$+$2^{nd}$) outperforms different Raw based versions by a margin ranging from 3.7\% to 12.91\% . 
These results indicate the efficacy of GSNE in modeling neighbourhood preference, which significantly improves price prediction performance.

\begin{figure}[ht]
\begin{center}
\includegraphics[width=0.7\textwidth]{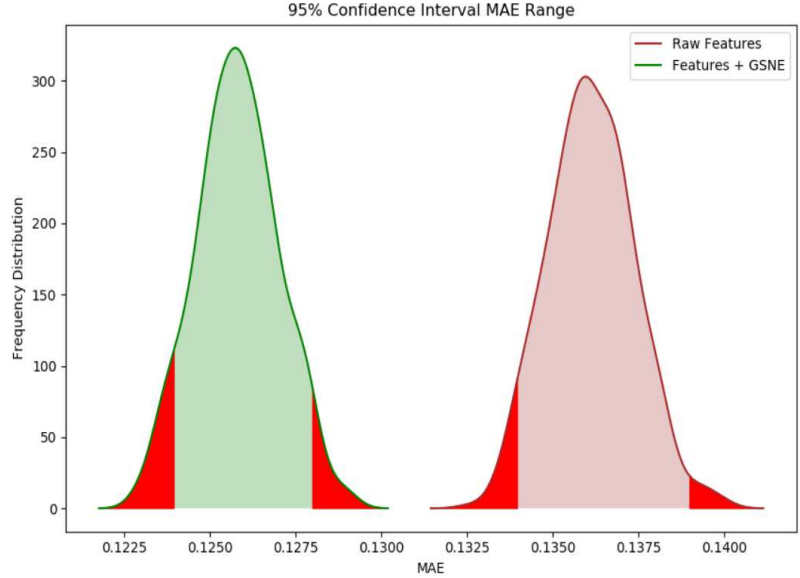}
\caption{Bootstrapped confidence interval comparison in Gradient Boost Regression. Here with \textit{95\%} confidence level, we can observe that the true MAE lies within \textit{0.124 and 0.128} when our spatial embedding is used with housing features, whereas for raw features only, this range lies within \textit{0.1338 and 0.139}.}
\label{fig:conf_int}
\end{center}
\end{figure}

\textbf{Confidence Interval:} We also calculate the 95\% confidence interval of MAE on both Raw and Raw+GSNE. 
For this comparison, we use Gradient Boosting Regression as our downstream model, since it gives the best result in Table \ref{tab:Price_Table}. 
From the result, we can observe that 95\% of the time the MAE lies within 0.124 and 0.128 for our embeddings (Raw+GSNE), whereas for Raw features only it lies within 0.1338 and 0.139. 
This analysis statistically validates the the efficacy of our proposed geo-spatial embedding in performance improvement.

\subsubsection{Varying Price Partitions}
\label{subsub:part}
To gain a better insight of the performance of our approach, we analyze the price prediction performance in different price quartiles of our dataset. 
We summarize these results in Table \ref{tab:Price_Partition}. 

From Table \ref{tab:Price_Partition} we observe that, in every price quartile, Raw+GSNE outperforms Raw. 
We also separately tested the performance on outliers: houses with prices outside of the $3\sigma$ range of the distribution, where $\sigma$ denotes the standard deviation of house price distribution. 
Outliers represent houses that are significantly more challenging to predict. 
Even on these outlier data, the features augmented with our embeddings outperform the Raw features only. 

From Table \ref{tab:Price_Partition}, we can also observe an interesting aspect of our geo-spatial embeddings. 
While the generated embeddings always improve prediction performance over Raw features, the improvement is even more pronounced in the $1^{st}$ and $4^{th}$ quartile price partitions. 
Intuitively, houses in the highest price category usually enjoy modern amenities in the neighborhood, whereas these facilities are somewhat limited for the cheapest houses. 
This explains why neighborhood context might be more influential in the two ends of price partitions.

\begin{table}[h]
\begin{center}
\caption{\label{tab:Price_Partition} Performance comparison in different price partitions using the Gradient Boosting regressor.}
\begin{tabular}{lcccccccc}
    \toprule
     & \multicolumn{2}{c}{MAE} &  \multicolumn{2}{c}{RMSE}\\
    \midrule
    Price Partition & Raw & Raw+GSNE  & Raw & Raw+GSNE\\
    $1^{st}$ Quartile & 0.119  & \textbf{0.110} & 0.163 & \textbf{0.155}  \\
    $2^{nd}$ Quartile & 0.106  & \textbf{0.102} & 0.138 & \textbf{0.133}  \\
    $3^{rd}$ Quartile & 0.122 & \textbf{0.115} & 0.157 & \textbf{0.149} \\
    $4^{th}$ Quartile & 0.173  & \textbf{0.159} & 0.228 & \textbf{0.212} \\
    Outside $3\sigma$ & 0.518 & \textbf{0.473} & 0.678 & \textbf{0.640} \\
    \bottomrule
\end{tabular}
\end{center}
\end{table}

\subsubsection{Impact of POIs on House Price}
\label{subsub:pois}
To investigate how different POIs impact house price prediction, we take the three types of POIs, i.e.\ Region, School, and Train Station, separately and train our embedding model on each of these nodes independently. 
To achieve this, we essentially take each bipartite partition (house-region, house-train, or house-school) separately and train GSNE on these networks. 
The generated embeddings are then channeled through the Gradient Boosting Regressor model. 
These results are presented in Table~\ref{tab:POI_Effect}. 

As we can observe from the table, even the consideration of each POI node separately gives us substantial performance improvements over raw features. 
Table~\ref{tab:POI_Effect} also reveals an interesting insights on how neighbouring transportation facilities and educational institutes are influential in predicting house prices. 
Nevertheless, training GSNE with all types of POIs yields the best results both in terms of MAE and RMSE. 

\begin{table}[h]
\centering
\caption{\label{tab:POI_Effect} Comparison of the effect of different POIs on house price with the Gradient Boosting Regressor.}
\begin{tabular}{lcccccccc}
    \toprule
     & MAE & RMSE\\
    \midrule
    Regions & 0.131  & 0.188 \\
    Train Station & 0.127 & 0.184 \\
    School & 0.126 & 0.182\\
    GSNE (All nodes) & \textbf{0.125} & \textbf{0.181}\\
    \bottomrule
\end{tabular}
\end{table}

\subsection{Visualization}
\label{subsection:vis}

\begin{figure*}[ht]
\begin{center}
\includegraphics[width=1\textwidth,height=0.2\textheight]{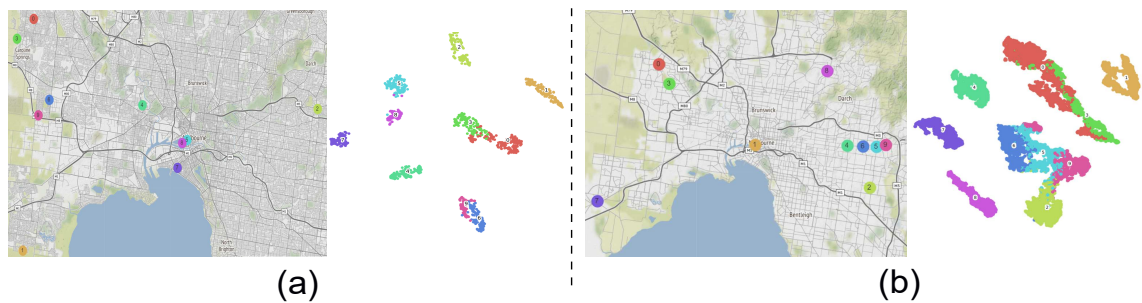}
\caption{2D visualization of the generated geo-spatial embeddings. (a) Visualization of houses adjacent to 10 regions (shown in map). These regions are taken as labels of different colors for the houses. (b) Visualization of houses adjacent to 10 train stations (shown in map). In both cases, we chose 10 of the POIs that have the highest number of houses in the neighbourhood for a better visibility.}
\label{fig:tsne}
\end{center}
\end{figure*}

We visualize how GSNE embeds neighbourhood information for different POI category to qualitatively analyse embedding quality.
In Figure~\ref{fig:tsne} we show separate t-SNE\cite{maaten2008visualizing} visualizations for the categories \emph{regions} and \emph{train stations}. 
Since each of these categories contains a large number of POIs, we select the top 10 POIs of each category that have the highest number of houses in its neighbourhood. 
We use these POIs as the labels of each house in its neighbourhood. 
In other words, if a house $i$ has an edge with a train station $j$, we use $j$ as the label of the house. 
Consequently, different colours in the visualization represent different POIs in that category.

Figure~\ref{fig:tsne}(a) contains the t-SNE visualization for the ten \textit{regions} as well as a map view of these regions, which shows their geographical locations. 
The t-SNE plots reveal that the separation is evident, such that houses in different regions are closely clustered to each other. 
A deeper look at the plot reveals more insights about the efficacy of GSNE. 
In the map we can see that region labeled as \textbf{0} and \textbf{3} are located in the top left corner (marked with red and green respectively), and they are close to each other. 
Their close vicinity is reflected in the embedding, as can be seen in the t-SNE plot, where the red and green clusters are very close to each other. 
It can also be observed that these two clusters overlap each other. 
With further investigation of the dataset, we discovered that the overlapping houses have a number of identical features, including median age, median house rent, median weekly income, and median rent. 
In other words, even though these two regions are not immediately next to each other geographically, they share highly similar features in terms of overall regional information. 

Another interesting case appears with region labeled as \textbf{5} and region \textbf{8}, coloured in cyan and purple respectively. 
In the map, we can see that these two regions are located close geographically. 
Yet they have very distinct feature sets. 
Houses in region \textbf{5} have almost twice the number of residents than region \textbf{8}, while only half of the median income of the residents of region \textbf{8}. 
Features such as median age, weekly house rent etc.\ are also distinct. 
These two cases give us an important insight about how GSNE embeds neighbourhood information in terms of the POI features. 
While it captures the neighbourhood information effectively, it also effectively captures the similarity of neighbourhoods in terms of their features. 
We see make similar observations about the other visible clusters. 
Regions labeled \textbf{1, 2, and 4} are very far from the other regions having distinctive features. 
They are also well separated from other clusters in the embedding space. 
Region \textbf{6 and 9} seem to be quite close geographically. 
Their feature sets are very similar with same median house rent, same median age, and similar median income. 
As can be observed, houses in these two regions are also closely clustered. 
From the analysis of this visualization, the efficacy of GSNE becomes quite evident.

In Figure \ref{fig:tsne}(b) we show the t-SNE visualization for \textit{train stations} and the corresponding map view. 
Here we can also observe good separations of clusters. 
From the map, it can be seen that the train stations labeled as \textbf{0} and \textbf{3} (marked red and green respectively) are located close to each other. 
These two stations also have identical feature sets. 
Their close vicinity can be observed in the t-SNE plot, showing the efficacy of our embedding model. 
Train station \textbf{1, 7, and 8} are far away from the other stations and each other, geographically, which is reflected in the t-SNE plot. 
On the other hand, station \textbf{2, 4, 5, 6, 9} are close geographically. 
In the embedding space, we see the clusters to be quite close for \textbf{2, 5, 6, 9}. 
However, the cluster for the station labeled as \textbf{4} is more separated from the above clusters. 
From the feature sets, we observed that the average time required to travel to other stations from station \textbf{4} is much lower compared to from stations \textbf{2, 5, 6, 9}, which is an indicator of better connectivity of station \textbf{4}. 
This essentially explains how the cluster of houses in the neighbourhood of station \textbf{4} achieves separation over the other nearby stations.

\subsection{Ablation Studies}

From the performance comparison in Table \ref{tab:Price_Table}, we can discern the effect of $1^{st}$- and $2^{nd}$-order proximities in the final result. 
We can observe that GSNE with either $1^{st}$- or $2^{nd}$-order proximity alone achieves noteworthy improvements over the performance of the raw features. 
Another notable fact is that, for regression models with higher expressive powers, we see comparable improvements for both $1^{st}$- and $2^{nd}$-order proximity version of GSNE. 
Nevertheless, in every case, the GSNE($1^{st}+2^{nd}$) sees the highest improvement. 
This indicates that the \textbf{local} and \textbf{global} features extracted respectively by $1^{st}$- and $2^{nd}$-order proximity complement each other, and their combination results in best performance in any chosen regression model. 
We also observe that the geo-spatial embedding alone (without concatenating with the raw housing features) achieves 0.222 MAE and 0.306 RMSE, which also validates our claims on the importance of neighbourhood contextual information in housing preferences.

The potency of GSNE embeddings in house price prediction is also visible from the confidence interval (C.I.) plot of the Gradient Boosting Regressor in Figure~\ref{fig:conf_int}. 
Here we bootstrap our dataset 500 times with 80\% train and 20\% test set, with which we train the Gradient Boost Regressor and examine its performance. 
From the plot, with 95\% probability we can observe that GSNE embeddings along with raw features achieves an MAE between 0.124 and 0.128. 
These two tails are respectively 7.32\% and 7.91\% better than the raw features confidence interval results where this range lies between 0.1338 and 0.139. 
These observations clearly demonstrate that the performance improvements of our approach is consistent over the whole dataset. 

From Table \ref{tab:POI_Effect}, we see that each of the different POI types gives different effects in house price prediction performance. 
Even though all of them improve performance when considered separately, we can see that the \emph{School} POI type improves the performance by the highest margin. 
This may indicate that in deciding the purchase of a house, buyers may value educational institutions over other POIs. We also observe that \emph{Train Station} POI type also gives a good house prediction performance boost as transport facilities in the neighborhood influence buyers' choices on a house. Yet, using all POI types in the embedding gives us the best performance, which indicates that all the POI types improve the performance of house price prediction.

\section{Conclusions}

In this paper, we have proposed a novel geo-spatial network embedding (GSNE) framework to accurately capture the geo-spatial neighborhood relationships between houses and surrounding POIs. The GSNE essentially learns low-dimensional Gaussian embeddings of nodes of a geo-spatial network. We have validated the efficacy of the GSNE  in the house price prediction task, where our detailed experimental evaluation shows that GSNE features combined with the raw housing features can predict house prices with a higher accuracy (i.e., 8.1\% lower MAE and 7.2\% lower RMSE) than that of the best performing state of the art methods that only consider standalone house features. It is important to note that though we validate the proposed GSNE on the house price prediction problem in this paper, our proposed geo-spatial embedding can be highly effective in answering other real estate queries like recommending similar houses, which is of independent interest. In future, we plan to explore how other complex house features such as textual description and images can be embedded in the multi-modal feature space to further enhance house price predictions.

\section{Acknowledgements}
We are grateful to Dr Zhifeng Bao, Associate Professor, RMIT University, Australia for sharing the Melbourne housing price dataset with us.

 \bibliographystyle{spmpsci}      
\bibliography{template.bib}


\end{document}